\documentclass[10pt, a4paper]{article}
\usepackage{lrec2022} 
\usepackage{multibib}
\newcites{languageresource}{Language Resources}
\usepackage{graphicx}
\usepackage{tabularx}
\usepackage{soul}
\usepackage{multirow}
\usepackage{booktabs}
\usepackage[table,xcdraw]{xcolor}
\usepackage[normalem]{ulem}
\useunder{\uline}{\ul}{}
\usepackage{titlesec}
\titleformat{\section}{\normalfont\large\bfseries\center}{\thesection.}{1em}{}
\titleformat{\subsection}{\normalfont\SmallTitleFont\bfseries\raggedright}{\thesubsection.}{1em}{}
\titleformat{\subsubsection}{\normalfont\normalsize\bfseries\raggedright}{\thesubsubsection.}{1em}{}
\renewcommand\thesection{\arabic{section}}
\renewcommand\thesubsection{\thesection.\arabic{subsection}}
\renewcommand\thesubsubsection{\thesubsection.\arabic{subsubsection}}

\usepackage{epstopdf}
\usepackage[utf8]{inputenc}

\usepackage{hyperref}
\usepackage{xstring}

\usepackage{color}

\title{Evaluation of HTR models without Ground Truth Material}

\name{Phillip Benjamin Ströbel\textsuperscript{1}, Simon Clematide\textsuperscript{1}, Martin Volk\textsuperscript{1}, Raphael Schwitter\textsuperscript{2}, \\
{\bf \large Tobias Hodel\textsuperscript{3}, David Schoch\textsuperscript{3}} }

\address{\textsuperscript{1}Department of Computational Linguistics, \textsuperscript{2}Department of Classics, \textsuperscript{3}Walter Benjamin Kolleg \\
         \textsuperscript{1,2}University of Zurich, \textsuperscript{3}University of Bern\\
         \{pstroebel, siclemat, volk\}@cl.uzh.ch\\
         raphael.schwitter@sglp.uzh.ch\\
         \{tobias.hodel, david.schoch\}@unibe.ch\\}

\abstract{
The evaluation of Handwritten Text Recognition (HTR) models during their development is straightforward: because HTR is a supervised problem, the usual data split into training, validation, and test data sets allows the evaluation of models in terms of accuracy or error rates. However, the evaluation process becomes tricky as soon as we switch from development to application. A compilation of a new (and forcibly smaller) ground truth (GT) from a sample of the data that we want to apply the model on and the subsequent evaluation of models thereon only provides hints about the quality of the recognised text, as do confidence scores (if available) the models return. Moreover, if we have several models at hand, we face a model selection problem since we want to obtain the best possible result during the application phase. This calls for GT-free metrics to select the best model, which is why we (re-)introduce and compare different metrics, from simple, lexicon-based to more elaborate ones using standard language models and masked language models (MLM). We show that MLM-based evaluation can compete with lexicon-based methods, with the advantage that large and multilingual transformers are readily available, thus making compiling lexical resources for other metrics superfluous.  
 \\ \newline  }

\begin{document}

\maketitleabstract

\section{Introduction}
\textit{Optical Character Recognition} (OCR) has become a well-established technique for digitising historical printed collections in libraries and archives. At the same time, \textit{Handwritten Text Recognition} (HTR) is also increasingly finding its way into these institutions \cite{terras2021inviting}. The digitisation efforts of libraries and archives simplify access to sought-after documents for researchers from various disciplines and allow them to pursue research questions they could either not or only very cumbersomely have answered without digital copies of the sources.

Still, we can only consider OCR and HTR to be ``research facilitators'' as long as they perform within reasonable accuracy ranges. Several studies have shown that inaccuracies in OCRed documents harm information retrieval and text mining techniques like named entity recognition and linking, topic modelling, and language modelling \cite{alex2014estimating,chiron2017impact,van_Strien_2020,hamdi2020assessing,pontes2019impact,hill2019quantifying}.

But what do we mean by ``reasonable accuracy ranges''? \newcite{holley2009good} shared the experiences of the \textit{National Library of Australia Newspaper Digitisation Program}, which found that OCR contractors and libraries usually agree that ``good'' OCR means $>98\%$, ``average'' OCR between 90\% and 98\%, and ``poor'' OCR any score  $<90\%$ character accuracy, respectively. \newcite{springmann2016automatic} defined ranges of $>95\%$ and between 90\% and 95\% character accuracy, but one time referring to the former as ``good'' and later as ``excellent'', and to the latter as ``reasonable'' and in another place as ``good'', respectively. Although opinions of what good OCR is differs, there is consensus that any score $<90\%$ means poor quality\footnote{In this paper, we mainly look at character error rates, which are simply the inverse of character accuracy. E.g., a 98\% character accuracy corresponds to a 2\% character error rate.}.

While \newcite{holley2009good} and \newcite{springmann2016automatic} were evaluating OCR for printed texts, these accuracy ranges naturally extend to text recognised from handwritten documents. However, HTR for personal handwriting styles is particularly challenging since a personal hand has much more variance than printed fonts or regular scripts like blackletter. Nonetheless, HTR has seen considerable improvements over the last decade, mainly thanks to the advent of neural networks \cite{graves2008offline,graves2009novel}. These developments fostered the creation of platforms like \textit{Transkribus}\footnote{\url{https://readcoop.eu/de/transkribus/}}, which facilitate the production of ground truth (GT), i.e., humanly transcribed material, and large-scale training of HTR models. Given suitable training data, the Transkribus-internal HTR+ model \cite{michael2018htr} can achieve character error rates (CERs) of about 5\% \cite{Mueh19}, therefore coming close to what \newcite{holley2009good} considered as ``good'' and \newcite{springmann2016automatic} already as ``excellent''.

Although HTR has made considerable progress, it has not yet found its way to a broader application in libraries and archives. A look at the digital platform for manuscript material for Swiss libraries and archives  \textit{e-manuscripta}\footnote{\url{https://www.e-manuscripta.ch/}} shows that in the category ``correspondence'' containing $\sim$45k titles, only 313, or $0.1\%$, contain transcriptions. With the output quality of modern HTR, libraries and archives can further process the scans that they accumulated over the years of mass digitisation and thus provide researchers and the public access to the content of manuscripts of various sorts.

However, the deployment of HTR models to produce this content comes with uncertainties. While we can evaluate the performance of HTR models during the development phase via accuracy or error scores on validation and test sets, we cannot do so during the application phase. Hence, there are only a few options to estimate the quality of the resulting texts:
\begin{enumerate}
    \item We compile a new GT from a representative sample of the new data.
    \item We rely on confidence scores.
    \item We find GT-free evaluation metrics.
\end{enumerate}
As concerns the first option, compiling a GT is costly, time-consuming, and requires human experts. Moreover, we suspect that we would only obtain a rough estimate of the overall quality \cite{hodel2021general}. As for the second option, confidence scores are not always available, and if they are, it is unclear whether and to what extent they correlate with actual accuracy scores across different documents and distinct models \cite{springmann2016automatic}. This leaves the third option as the only remaining one, which we will address in this paper. We propose different metrics in Section \ref{metrics} and investigate whether there is a significant correlation between the introduced metrics and the actual CER. 

The scrutiny of metrics for performance evaluation has practical implications for libraries and archives: on the one hand, they can provide an estimate of the text quality, which is relevant for researchers working with this data. On the other hand, GT-free metrics allow for model selection when several models have produced hypotheses. Factors like the number of authors in a particular collection, the size of the GT, the skew of the author distribution, scan quality, and the like justify the training and testing of multiple models with different parameters and training data. In order to produce the best possible text during the application phase, it is necessary to apply all models and decide which model produced the best result for each document. In such a case, GT-free evaluation metrics can assist libraries and archives in making this choice. Hence, we also test whether GT-free evaluation metrics are suitable for model selection.

To summarise, in this paper, we investigate a collection of GT-free evaluation metrics to provide a quality estimation of HTR model results. Moreover, we examine whether we can use the metrics to select the best performing models.

\section{Related Work}
\label{lit}
Studies examining the evaluation of HTR output are few, which might be because baselines on popular data sets like the IAM \cite{marti2002iam} database for handwritten English sentences are indicated in standard measures (CERs or word error rates) \cite{sanchez2019set}. \newcite{neudecker21} pointed out that although the evaluation of OCRed output is standardised by measures like CERs, different implementations in tools that assist the evaluation make it difficult to compare model performances. Additionally, \newcite{schambach2010reviewing} stated that such standard measures to rank models ``may be of limited significance for the decision to choose a recognition system for a real-life application.'' His work showed that for the recognition of postal addresses, the final performance metric and, e.g., error limitation, i.e., the rejection of uncertain results, for different model architectures do not necessarily correlate.

Whenever confidence scores are available, we can exploit them to estimate the text quality. For example, \newcite{sarkar2001triage} used confidence score outputs by an OCR model to build a triage system for OCRed documents. They classified documents to bypass manual inspection, which they successfully achieved for 41\% of all documents in their validation set. Overall, they could show that their triage method significantly sped up the document verification process.

\newcite{clausner_quality_2016} suggested an OCR quality prediction system that assesses the accuracy of OCR results by any given system. Their classifier relies on 28 features, including metadata, image and layout information, textual features, and confidence scores. Their model was trained on newspaper data and provided reliable OCR quality estimates with an average error rate of 6.1\% on a bag-of-words basis, indicating suitability for OCR quality estimation. In addition, a more detailed feature analysis of their classifier showed that the confidence scores are the third most important feature, only topped by lexical features. 

Work by \newcite{springmann2016automatic} on the automatic OCR quality estimation of historical printings also showed that the relationship between accuracy and confidence scores is significant (in contrast to \newcite{schambach2010reviewing}).\footnote{We believe that the extent of how much we can trust a model's confidence scores strongly depends on the model itself.} Their approach investigated two measures: a mean token lexicality score based on the text-channel-model by \newcite{reffle2013unsupervised} and confidence values that the OCR model assigns to its output characters. Moreover, they further improved their ``standard'' models that they had already trained by continuous training with different amounts of ``pseudo ground truths'' consisting of text output where standard models return high confidence values. Lastly, \newcite{springmann2016automatic} used the confidence values in order to rank the models during training, which enabled them to choose the best one according to the confidence values after all training runs had been completed. 

\newcite{alex2014estimating} illustrated the usefulness of lexicality in OCR outputs and advocated for a much simpler evaluation procedure. They examined the influence of OCR errors on named entity recognition in historical documents and found that especially recall is affected. They suggested measuring the quality of a text by simply computing the ratio of all words in a document and words occurring in an English dictionary. The test of their metric against human evaluation showed a correlation between their score and human judgements.

\newcite{salah2015ocr} developed an approach that operates without external language resources. Their method relied on two OCR outputs: an OCR result and a second OCR output by another system called the reference (i.e., an uncertain GT). They used word-based character agreement rates and mutual information scores as features for a Support Vector Regressor that predicts the word recognition rate. They found their approach reliable, with an R² score of 0.95.

We can also use error classifiers in OCRed texts as quality estimators. \newcite{jatowt2019post} proposed an error detector that uses several word- and character-n-gram measures as features to train an OCR error detector. They also introduced a so-called \textit{peculiarity index} which they defined as the root-mean-square of the indices of its trigrams. The higher the peculiarity index of a word, the more unusual its trigram composition is and, consequently, the more unlikely it is to be correct. Their error detector works best with all features included, reaching an F-score of 79\% on monographs and 70\% on periodicals.

\section{Method}
\begin{figure*}[!h]
    \centering
    \includegraphics[scale=0.4]{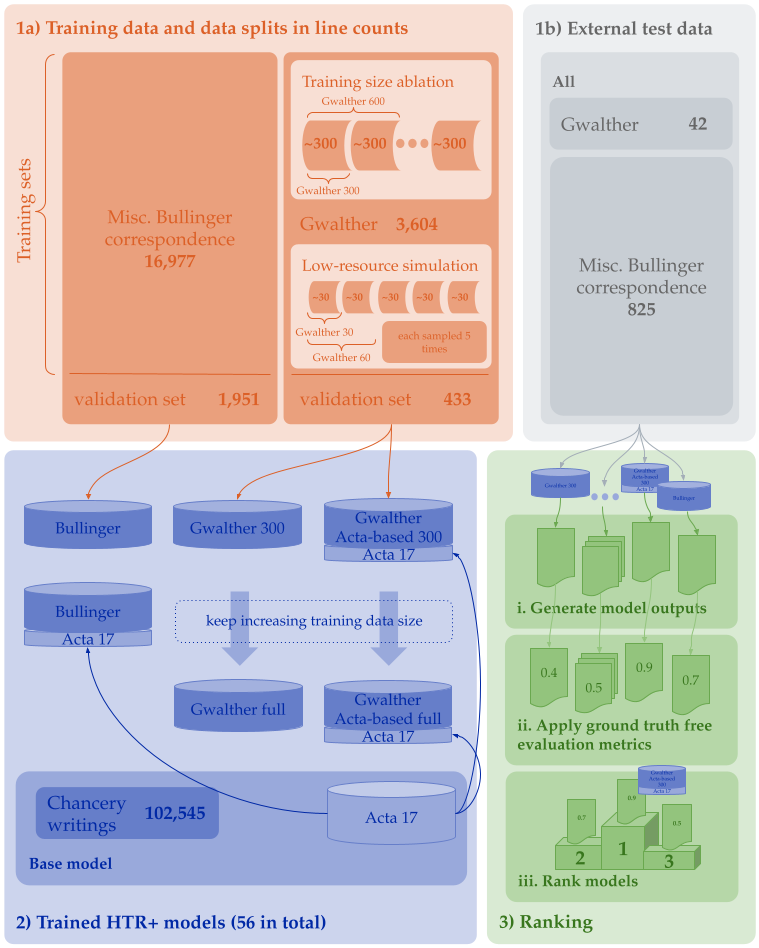}
    \caption{Training and evaluation procedure.}
    \label{fig:data}
\end{figure*}

Our goal is to evaluate different GT-free metrics and to check for each of them if it is suitable to rank HTR models. In the first step, we train many models on data sets we specify in more detail in Section \ref{data}. Next, we apply all models to different test sets and compute all metrics introduced in this Section. This procedure results in a ranking for each metric. At the same time, we can rank all the models we have trained according to the CER they achieve on the test set, which we use as our reference ranking. Ideally, a metric's ranking would be the same as the ranking according to the CER. We use Spearman's rank correlation measure to learn whether the reference ranking correlates with the metrics' rankings. Its coefficients tell us at what significance level each two ranking pairs correlate. Fig.~\ref{fig:data} summarises the pipeline, which we will describe in more detail in this and the forthcoming sections.

\subsection{
Metrics
}
\label{metrics}
The code and language resources to run the metrics will be made available with the publication of the conference proceedings.\footnote{\url{https://github.com/pstroe/atr-eval}}

\subsubsection{Lexicon-based Metrics}
\newcite{clausner_quality_2016} showed that lexical features are most suitable to predict OCR quality, while \newcite{alex2014estimating} indicated that a simple ratio of recognised words against a lexicon correlates well with human judgement about OCR quality. We thus include lexicon-based methods in our metrics.
\paragraph{Token Ratio}
The token ratio determines the percentage of tokens recognised by the HTR model that also occur in a reference. Let $T$ be the tokens in the HTR result and $\mathcal{V}$ the corresponding vocabulary. The token ratio is given by

\begin{equation}
\label{eq:tokenratio}
\mathcal{R}_{\mathrm{token}}=\frac{c(T \in \mathcal{V})}{c(T)}.
\end{equation}

This measure performs a lexicon lookup, where the lexicon is based on a collection of texts in the same language.


\paragraph{Character \textit{N}-gram Ratio}
\newcite{jatowt2019post} included an \textit{n}-gram analysis in the OCR error detection process. We also include character \textit{n}-gram ratios against a lexicon, but without the peculiarity index that \newcite{jatowt2019post} introduced.

We consider the ratios of $2$- to $7$-grams. For each \textit{n}-gram variant, we build the corresponding \textit{n}-gram vocabulary from the reference text and the HTR results. Be $\mathcal{G}$ the set of \textit{n}-grams in the reference text and $N$ the \textit{n}-grams in the HTR result. The \textit{n}-gram ratio is given by
\begin{equation}
\label{eq:ngramtokenratio}
\mathcal{R}_{n\mathrm{-gram}}=\frac{c(N \in \mathcal{G})}{c(N)}.
\end{equation}

\subsubsection{Language Modeling Perplexity}
RNN-based HTR architectures learn an implicit character language model (LM) \cite{sabir2017implicit}. Other approaches pre-train a LM and incorporate it in an HTR model to boost performance \cite{kang2021candidate}. LMs thus form an integral part of the HTR process, so we suggest using external ones to evaluate HTR results. This is the first approach that uses LMs for HTR quality estimation to the best of our knowledge. We differentiate between statistical LMs and transformer-based LMs.

\paragraph{Statistical Language Modeling Perplexity}
The task of LMs is to predict a word given a sequence of preceding words.  Probabilistic LMs learn from massive text corpora which  words must follow each other. A statistical LM is given by
\begin{equation}
\label{eq:mle}
P_{MLE}(w_n|w_1 ... w_{n-1})=\frac{c(w_1 ... w_n)}{c(w_1 ... w_{n-1})}
\end{equation}
where \textit{MLE} refers to \textit{Maximum Likelihood Estimation} and \textit{n} to the order of \textit{n}-grams. For example, a bi-gram LM would predict the next word based on the preceding word.

We can evaluate LMs with the perplexity (PPL) score, a measure from information theory. For example, given a LM $P$ and test set $W$ consisting of words in their sentential contexts, the word-level PPL measures the surprisal of the model when computing the likelihood of the sequence of words: 
\begin{equation}
\label{eq:ppl}
PPL(W) = 2^{H(W)}
\end{equation}

where $H$ is the cross-entropy of the sequence $W$:

\begin{equation}
\label{eq:crossentropy}
H(W) = -\frac{1}{N} \sum_{i=1}^{N} \mathrm{log}_2P(w_i)
\end{equation}

The lower the PPL score, the less surprised the model is, and hence, the better the text quality should be for HTR results.

\paragraph{Pseudo-Perplexity}
With the advent of deep learning, language modelling has considerably increased accuracy, not least thanks to transformer-based \cite{vaswani2017attention} methods like, e.g., BERT \cite{devlin2018bert}. BERT is an approach that uses the transformer blocks and attention heads and learns two tasks: (1) the prediction of tokens that are masked in the input and (2) the prediction of whether the next sentence is a natural continuation of the input sentence. As such, the authors designed BERT as a language representation model suitable to be fine-tuned on many tasks like natural language understanding, question answering, or grounded common sense inference. There is a large amount of research striving to improve and modify \textit{masked language models} (MLMs) like BERT. One modification we will use is RoBERTa \cite{liu2019roberta}, which used longer training over more and larger batches and more data, the omission of the next sentence prediction, and other tweaks to make BERT more robust.

Whilst PPL is a suitable metric for traditional LMs, the MLM objective does not permit a direct calculation of PPL. Instead, \newcite{salazar2019masked} proposed the use of \textit{pseudo-perplexities} (PPPLs). The PPPL is the exponential of the word-normalised sums of each token's conditional log probability in all output sentences to evaluate. More concretely, given an input sentence, we mask each token once, add up the conditional log probabilities of all tokens and normalise by the number of words in the sentence. We use the MLM scorer\footnote{\url{https://github.com/awslabs/mlm-scoring}} framework proposed by the authors for our experiments.

\section{Data and Models}
\label{data}

\subsection{HTR Data}
In contrast to the evaluation methods in Section \ref{lit}, we work with historical handwritten data, more concretely, with samples from Heinrich Bullinger's (1504-1575) extensive correspondence. The Zurich State Archives and the Zurich Central Library have preserved some 2,000 letters that Bullinger wrote and 10,000 letters that he received. 80\% of the letters are Latin, most of the others in Early New High German. About 3,000 letters had already been manually transcribed, edited, and published in printed form in efforts that spanned three decades.\footnote{They are available on \url{http://teoirgsed.uzh.ch/}.} Another 5,700 letters have been transcribed by various scholars and are available as electronic texts, albeit in uncertain quality. In our ongoing project \textit{Bullinger digital}\footnote{\url{https://www.bullinger-digital.ch/}}, we make the scans and transcriptions accessible for researchers and the public. Moreover, we will produce automatic transcriptions for the roughly 3,000 letters for which no transcriptions exist and thus will face model selection and HTR quality estimation issues.

\begin{figure}[t]
    \centering
    \includegraphics[scale=0.38]{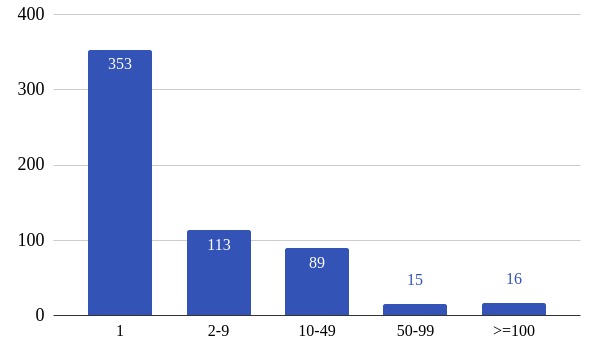}
    \caption{Author distribution of the Bullinger correspondence. Number of letters on the x-axis.}
    \label{fig:authorhist}
\end{figure}

\begin{figure}[t]
    \centering
    \includegraphics[scale=0.22]{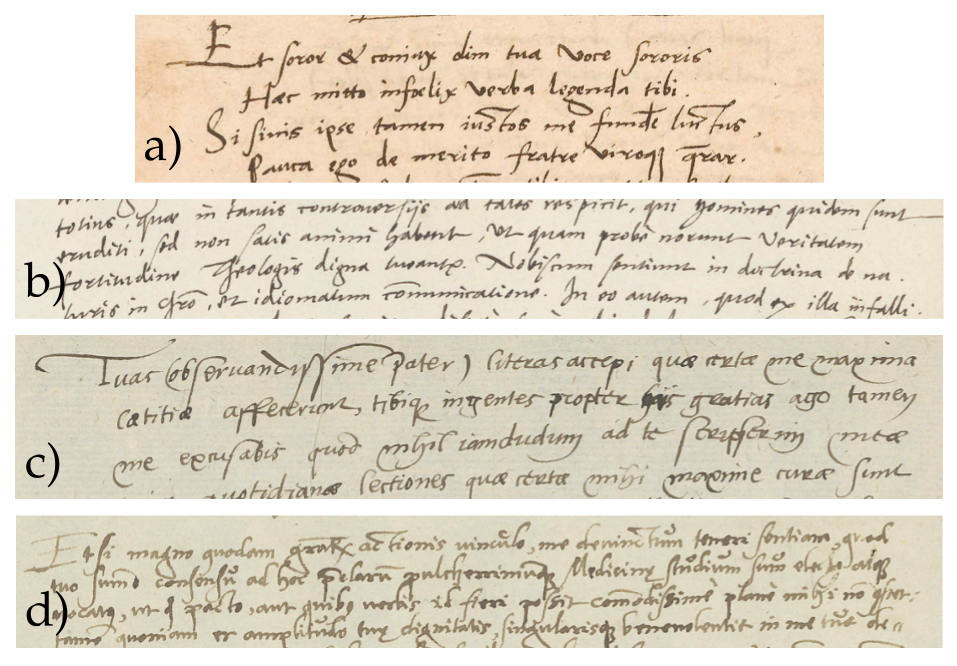}
    \caption{Different handwriting styles. a) poetry by Rudolf Gwalther, letters by b) Rudolf Gwalther, c) Matthieu Coignet, and d) Kaspar Wolf.}
    \label{fig:hwstyles}
\end{figure}

Bullinger received many more letters than he wrote, so the author distribution is heavily skewed. Fig.~\ref{fig:authorhist} exemplifies this fact; over 350 contemporaries only ever wrote once to Bullinger. This skew presents a considerable hurdle for HTR models since the presence of many different hands makes it more difficult to generalise. Fig.~\ref{fig:hwstyles} shows examples of how the handwritings differ among authors. Especially in such a setting, it can be necessary to train multiple models with different author distributions and choose the model that performs best.

Our training data for these models comes from two sources. Firstly, we found that a correspondent of Bullinger's, Rudolf Gwalther, wrote a volume called \textit{Lateinische Gedichte}, whose scan images and partial transcriptions are available on \textit{e-manuscripta}\footnote{\url{https://www.e-manuscripta.ch/zuz/content/titleinfo/1111284}}. We loaded the images into the Transkribus interface, applied layout recognition to identify lines, and aligned the transcriptions with the lines. We made minor corrections during this process, mainly concerning inconsistent capitalisation and punctuation. The resulting ``Gwalther'' data set consists of $\sim$4,000 lines and is one of our training sets (see box 1a on the right in Fig.~\ref{fig:data}).\footnote{This set is publicly available on \url{https://doi.org/10.5281/zenodo.4780947}.}

Secondly, we aligned the scan images that our suppliers have already delivered with the already published transcriptions by applying Transkribus' Text2Image module, which resulted in an automatically aligned GT of roughly 20,000 lines that we call ``Misc.~Bullinger correspondence'' (see box 1a on the left in Fig.~\ref{fig:data}). An analysis of the author distribution shows a heavily skewed picture with 69 different authors, where about 4,600 lines stem from Oswald Myconius and only two, e.g., from Georg Cassander.

We also compiled two separate test sets. On the one hand, we sampled 825 lines from a portion of scans without transcriptions (box 1b in Fig.~\ref{fig:data}) and manually transcribed the lines. Again, the author distribution is heavily skewed. Moreover, 35\% of the 825 lines are by authors, which we also find in the ``Misc.~Bullinger correspondence''. On the other hand, we include a separate 57 lines letter from Gwalther that is not part of any training or test data. We observed that Gwalther's writing style differs in the poetry volume and the letters, as Fig.~\ref{fig:hwstyles} shows. Models trained on ``Gwalther'' data should nevertheless perform better on this test set than other models.

\subsection{Language Resources}
\label{langres}
Since we worked with Latin correspondence, we collected Latin reference material (i.e., we did not use texts in Early New High German). For the experiments in Section \ref{experiment}, we downloaded the Latin part of the CC-100 corpus\footnote{\url{http://data.statmt.org/cc-100/}, accessed: 29.09.2021.} \cite{conneau2019unsupervised,wenzek2019ccnet}. The raw file contains 2.5G of text, i.e., 390 million tokens. Next, we removed generic "Lorem ipsum" text, deduplicated the corpus and retained only lines consisting of characters in the Latin alphabet, numbers, and a selection of punctuation marks. Finally, we tokenised and normalised the corpus with the help of the \textit{Classical Language Toolkit} CLTK \cite{johnson-etal-2021-classical} and split it into sentences. After preprocessing $\sim$206 million tokens remain.

\subsection{Models}
\subsubsection{HTR Models}
\begin{table*}[!htbp]
\centering
\resizebox{0.99\textwidth}{!}{
\begin{tabular}{lcrrrrrrrrrrrrrr}
\toprule
                              & \multicolumn{15}{c}{\textbf{Number of lines for training}}                                                                    \\ \midrule
 &
  \multicolumn{1}{r}{30} &
  90 &
  150 &
  300 &
  600 &
  900 &
  1200 &
  1500 &
  1800 &
  2100 &
  2400 &
  2700 &
  3000 &
  3300 &
  3600 \\ \midrule
No base              & \multicolumn{1}{r}{39.33} & 14.37 & 11.39 & 7.28 & 5.55 & 4.96 & 4.36 & 4.14 & 3.9  & 3.94 & 3.64 & 3.59 & 3.36 & 3.24 & 3.29 \\
Acta-based           & \multicolumn{1}{r}{13.67} & 8.38  & 7.08  & 5.5  & 4.42 & 4.03 & 3.66 & 3.43 & 3.27 & 3.13 & 3.09 & 3.2  & 2.82 & 2.8  & 2.74 \\ \toprule
                     & \multicolumn{15}{c}{\textbf{Other models (evaluated on the ``Gwalther'' validation set)}}                                                                                    \\ \midrule
Bullinger    & \multicolumn{15}{c}{6.99}                                                                                                     \\
Bullinger+Acta & \multicolumn{15}{c}{6.56}                                                                                                     \\
Acta\_17             & \multicolumn{15}{c}{14.66}                                                                                                    \\
Spruchakten & \multicolumn{15}{c}{15.95}                                                                                                    \\ \bottomrule
\end{tabular}
}
\caption{CERs for all models trained on different training data and number of lines on their respective test sets.}
\label{modelcomp}
\end{table*}
We trained all our HTR models with Transkribus and the HTR+ model architecture for 50 epochs. For both training data sets in box 1a in Fig.~\ref{fig:data}, we set aside 10\% of the corpus for validation and 90\% for training. Transkribus offers the possibility to include base models in the training process. We selected the publicly available model ``Acta\_17 HTR+''.\footnote{The data (chancery writings) and model were compiled by Alvermann et al.~from the University of Greifswald. Period: 1580-1705, size: $\sim$600k words in Latin and Low German from 1k different authors, epochs: 1k.} A base model initialises the model’s weights and allows for fine-tuning the model on novel data. This way, the model knows something about handwriting before seeing the new training data, leading to faster convergence and better performance. We trained a variant with and without a base model for every data set.

For the ``Gwalther'' data set, we additionally conducted a size ablation, starting at 300 lines (corresponding to roughly ten pages) up to the usage of the entire training set in increments of 300 lines, following a similar approach as detailed in \newcite{strobel-etal-2020-much}. Moreover, we simulated an extreme low-resource setting, in which we sampled 30, 90, and 150 lines five times and used those as training data.

We sketched the training procedure in box 2 in Fig.~\ref{fig:data}. This training regime results in a total of 56 models. We only considered the best performing model per training set size for the low-resource simulation setting, thus discarding 24 models. We also evaluated the base model Acta\_17 HTR+ and the \textit{Spruchakten\_M\_2\-11} model on the ``Gwalther'' validation set and the test sets.\footnote{The model was trained on a subset of the \textit{Acta\_17} training set: period: 1583-1653, size: 246k words, epochs: 1k.} The reason for including this model is to check whether the effects of applying a model on paleographically more related data (since it's closer to Bullinger's time) are more pronounced than just corpus size.

We present the results of all the models in Table \ref{modelcomp}. We note a considerable improvement of performance the more data we add, while models trained with a base model consistently perform better than their counterparts without a base model. The Spruchakten model, although trained on data closer to Bullinger's time of living, performs worse than the Acta\_17 model, which indicates that more data helps.

\subsubsection{Language Models}
We used the Latin corpus presented in Section \ref{langres} to train all LMs. We applied a data split of 90\% for training and 10\% for testing and kept it constant for all models.
\paragraph{Statistical LM}
We used the \textit{KenLM} package \cite{heafield2011kenlm,heafield2013scalable} to compute a 5-gram Kneser-Ney interpolated LM on the Latin training data. It achieves a PPL score of 23.13 on the Latin corpus test set.
\paragraph{Masked Language Models}
For the evaluation with the help of MLMs, we used the pre-trained multilingual BERT model based on \newcite{devlin2018bert}. Its training data consists of Wikipedia articles in 104 different languages, among which we also find Latin. It is available via the Hugging Face model hub \cite{wolf2019huggingface}.\footnote{\url{https://huggingface.co/bert-base-multilingual-cased}}

Since the multilingual BERT model uses WordPiece tokenisation and a shared vocabulary of 110k subtokens, we hypothesise that a single-language MLM should be superior to multilingual BERT. Thus, we trained a RoBERTa model with standard settings on the training set of our Latin corpus. We make the model\footnote{\url{https://huggingface.co/pstroe/roberta-base-latin-cased}}, as well as the training and test data\footnote{\url{https://huggingface.co/datasets/pstroe/cc100-latin}} available on the Hugging Face hub.  

\section{Experiments}
\label{experiment}
\subsection{HTR Quality Estimation Suitability}
\label{htrqualityest}
After training the HTR and language models and preparing our language resource, we want to investigate whether the metrics introduced in Section \ref{metrics} are suitable for HTR quality evaluation. We aim to establish a correlation between the CERs and the different metrics. Thus, we applied each model to our test set and computed the scores of our metrics.  For evaluating PPPL, PPL, and token ratio, we preprocessed the HTR result in the same way as our Latin reference corpus to guarantee a fair comparison.

Our null hypothesis H\textsubscript{0} is that the metrics' scores on the test data and the CERs that the models achieve on the same data do not correlate. We follow \newcite{springmann2016automatic}'s approach in fitting linear models with the CERs as dependent variables and the scores of the metrics as predictors using the R package \cite{r}. The adjusted R² informs us about the goodness of fit, where a higher value indicates the model can fit the data well. This procedure helps us determine whether a metric reliably estimates HTR quality.

\subsection{Ranking Ability}
The ranking tells us whether our metrics capture differences in HTR models. Our null hypothesis H\textsubscript{0} states that there is no correlation between the rankings of different models based on CERs and the rankings based on the metrics. Rejecting H\textsubscript{0} would prove the existence of such a correlation and validate the proposed metrics. Furthermore, the strength of the correlation between the CER and each of the metrics allows us to determine which metric the ranking should be based on.

We compute Spearman's rank correlation coefficient $\rho$ between the reference and metric rankings to check for correlations. It is given by 
\begin{equation}
\label{eq:spearman}
    \rho=1-\frac{6\sum d_{i}^2}{n(n^2-1)},
\end{equation}
where $d_i$ denotes the difference between two ranks and $n$ is the number of ranks. A comparison against critical values indicates whether the correlation is significant, thus rejecting H\textsubscript{0}.

Applied to our data, this means that we ran all our models on the Gwalther\textsubscript{57} and the Bullinger\textsubscript{825} test sets in block 1b in Fig.~\ref{fig:data}, computed the corresponding CERs, and ranked the models according to the CERs. These are our reference rankings. We added the reference ranking for the Gwalther\textsubscript{433} validation set in block 1a. Table \ref{modelcomp} shows that many models perform in similar ranges, especially for the Acta-based models. Our metrics should reflect this by returning higher values when we apply the models to the validation set. We then took each model output for the Gwalther\textsubscript{433}, Gwalther\textsubscript{57}, and the Bullinger\textsubscript{825} test set and applied our metrics to it. This produces a score for each model and metric, which we used to rank the models. Lastly, we computed Spearman's $\rho$ by comparing each metric ranking to its corresponding reference ranking. For example, we compared the model ranking of the PPL metric on the Gwalther\textsubscript{57} test set against the model ranking according to the CERs on the same set.

\subsection{Hit Top-N}
The last investigation we conducted is a check for each metric on how well it identifies a model in the Top-1, Top-3, or Top-5. As we see in Table \ref{modelcomp}, the Top-5 Acta-based models have a difference of at most 0.35 percentage points, which lead us to assume that for practical applications, a method that would identify either of those five models as the best performing model was suitable for model selection. A single factor ANOVA test on the Gwalther\textsubscript{433} with a p-value of 0.71 hardens our suspicions that the means between the Top-5 performing models are not significantly different. 

\section{Results and Discussion}
\begin{figure}[h]
    \centering
    \includegraphics[scale=0.15]{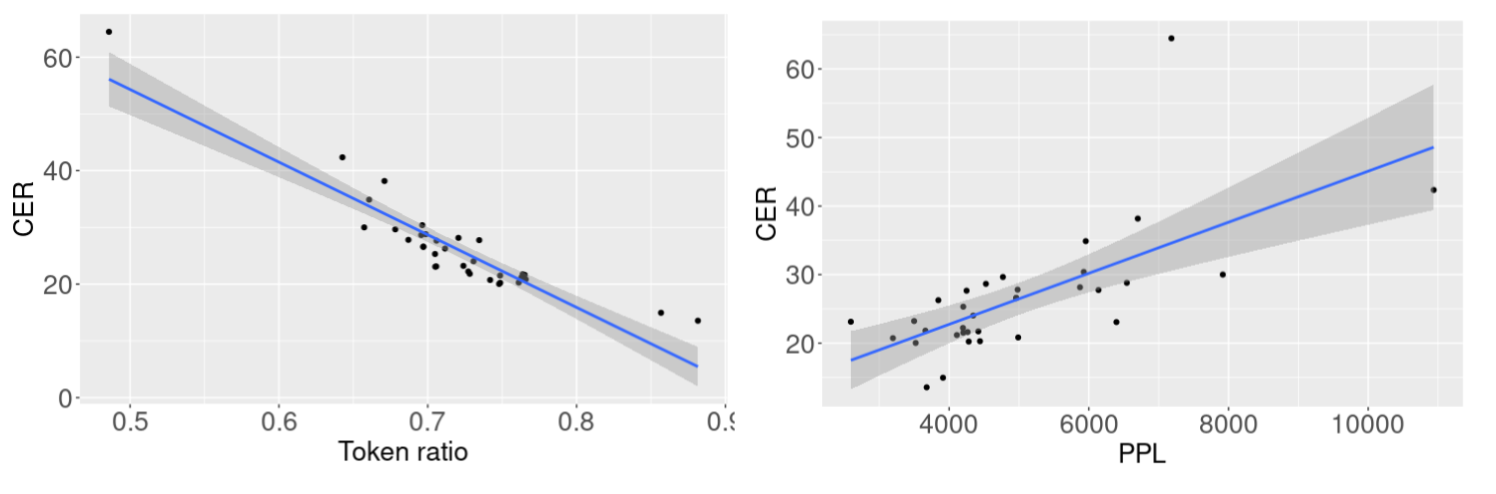}
    \caption{Model fit for two measures (Token ratio left and PPL right) for the Gwalther\textsubscript{57} test set. Regression line in blue, 95\% confidence interval in grey.}
    \label{fig:modelfit}
\end{figure}

Table \ref{r2} shows the adjusted R² values for the models we fitted for each CER vs.~metric pair. The superscript behind each score indicates the polynomial degree of the model. The higher the R² score, the better the model fits the data, i.e., the  higher the correlation between metric and actual CER. We exemplify this with two metrics for the Gwalther\textsubscript{57} test set from the Table \ref{r2} in Fig.~\ref{fig:modelfit}. Each black dot represents a model. We notice the linear relationship (blue line) between metric and CER.

The big picture from HTR quality estimation suitability confirms that our metrics generally find it easier to score the Gwalther\textsubscript{433} set. This is only logical since we trained most of our models on the ``Gwalther'' set.

\begin{table}[ht]
\centering
\resizebox{0.5\textwidth}{!}{
\begin{tabular}{llrrr}
\toprule
   &                              & \multicolumn{3}{c}{\textbf{Adj. R²}}                             \\
\multicolumn{2}{l}{\multirow{-2}{*}{\textbf{Metrics}}} &
  \multicolumn{1}{p{1.5cm}}{Gwalther\textsubscript{433}} &
  \multicolumn{1}{p{1.5cm}}{Bullinger\textsubscript{825}} &
  \multicolumn{1}{p{1.5cm}}{Gwalther\textsubscript{57}} \\ \midrule
& \multicolumn{1}{l|}{BERT}    & \multicolumn{1}{r|}{0.99²}  & \multicolumn{1}{r|}{0.80²} & 0.92² \\
\multirow{-2}{*}{PPPL} &
\multicolumn{1}{l|}{RoBERTa} & \multicolumn{1}{r|}{0.98¹}  & \multicolumn{1}{r|}{0.69²} & 0.88² \\ \midrule
\multicolumn{1}{l}{PPL}  & \multicolumn{1}{l|}{Statistical LM}        & \multicolumn{1}{r|}{0.65²}  & \multicolumn{1}{r|}{0.52¹} & 0.42¹ \\ \midrule
\multicolumn{2}{l|}{Token ratio}  & \multicolumn{1}{r|}{0.99¹}  & \multicolumn{1}{r|}{0.95\textsuperscript{4} } & 0.92³ \\ \midrule
\multirow{6}{*}{Character \textit{n}-grams} &
  \multicolumn{1}{r|}{2-gram} &
  \multicolumn{1}{r|}{0.42²} &
  \multicolumn{1}{r|}{-0.01¹} &
  0.14² \\
   & \multicolumn{1}{r|}{3-gram}  & \multicolumn{1}{r|}{-0.03¹} & \multicolumn{1}{r|}{0.09¹} & -0.03¹ \\
   & \multicolumn{1}{r|}{4-gram}  & \multicolumn{1}{r|}{0.80²}   & \multicolumn{1}{r|}{0.35¹} & 0.06¹ \\
   & \multicolumn{1}{r|}{5-gram}  & \multicolumn{1}{r|}{0.96²}  & \multicolumn{1}{r|}{0.79¹} & 0.81¹ \\
   & \multicolumn{1}{r|}{6-gram}  & \multicolumn{1}{r|}{0.99²}  & \multicolumn{1}{r|}{0.96³} & 0.96² \\
   & \multicolumn{1}{r|}{7-gram}  & \multicolumn{1}{r|}{0.99²}  & \multicolumn{1}{r|}{0.99²} & 0.97² \\ \bottomrule
\end{tabular}
}
\caption{R² values for all metrics compared to the performance of the models on the respective validation or test sets. The superscript indicates the polynomial degree of the model we fitted to the data.}
\label{r2}
\end{table}

To put our results in relation to previous work, we would like to point out that some of our R² scores are considerably higher than what \newcite{springmann2016automatic} reported for their confidence vs.~CER and lexicality vs.~CER correlations. However, our results confirm the importance of the lexicality of the output and that its measuring is suitable for HTR quality estimation. As concerns character \textit{n}-grams, the general tendency is, the higher the order of the \textit{n}-gram, the better it correlates with the CER.

As for the (P)PPL metrics, especially the scoring with the multilingual BERT model results in reliable estimates for HTR quality. We found it surprising that a dedicated Latin model does not perform better than multilingual BERT. However, the upside of this finding is that we can use a pre-trained model out of the box for HTR quality estimation without having to train additional models. Moreover, the solid performance of PPPL makes the collection of language-specific data superfluous.

Moving on to the examination of the ranking capabilities of our metrics, we find that most metrics' rankings exhibit strong correlations to the reference rankings, as Table \ref{tab:spearman} shows. Overall, the ranking correlation shows a similar picture as the R² values do. Again, the token ratio metric shows strong correlations. The PPL of the statistical LM cannot keep up with the other measures, while our newly proposed application of PPPL for ranking purposes perform equally well as the token ratio metric, at least on the Gwalther data.

\begin{table}[]
\centering
\scalebox{0.65}{
\begin{tabular}{lr|rrr}
\toprule
\multicolumn{2}{l}{} &
  \multicolumn{3}{c}{\textbf{Ranking Reference (CER)}} \\
\multicolumn{2}{l}{\multirow{-2}{*}{\textbf{Metrics}}} &
  Gwalther\textsubscript{433} &
  Bullinger\textsubscript{825} &
  Gwalther\textsubscript{57} \\ \midrule
 &
  \multicolumn{1}{l|}{BERT} &
  {\color[HTML]{e38052} {\ul \textit{0.98}}} &
  {\color[HTML]{e38052} {\ul \textit{0.90}}} &
  {\color[HTML]{e38052} {\ul \textit{0.90}}} \\
\multirow{-2}{*}{PPPL} &
  \multicolumn{1}{l|}{RoBERTa} &
  {\color[HTML]{e38052} {\ul \textit{0.96}}} &
  {\color[HTML]{e38052} {\ul \textit{0.82}}} &
  {\color[HTML]{e38052} {\ul \textit{0.85}}} \\ \midrule
PPL &
  \multicolumn{1}{l|}{Statistical LM} &
  {\color[HTML]{e38052} {\ul \textit{0.78}}} &
  {\color[HTML]{e38052} {\ul \textit{0.65}}} &
  {\color[HTML]{e38052} {\ul \textit{0.71}}} \\ \midrule
Token ratio &
   &
  {\color[HTML]{e38052} {\ul \textit{0.98}}} &
  {\color[HTML]{e38052} {\ul \textit{0.95}}} &
  {\color[HTML]{e38052} {\ul \textit{0.91}}} \\ \midrule
  \multirow{6}{*}{Character \textit{n}-grams}
 &
   
  2-gram &
  -0.28 &
  0.28 &
  0.13 \\
 &
   
  3-gram &
  0.01 &
  {\color[HTML]{3353b7} {\ul \textit{0.48}}} &
  {\ul \textit{0.36}} \\
 &
   
  4-gram &
  {\color[HTML]{e38052} {\ul \textit{0.58}}} &
  {\color[HTML]{e38052} {\ul \textit{0.62}}} &
  0.15 \\
 &
   
  5-gram &
  {\color[HTML]{e38052} {\ul \textit{0.90}}} &
  {\color[HTML]{e38052} {\ul \textit{0.88}}} &
  {\color[HTML]{e38052} {\ul \textit{0.70}}} \\
 &
   
  6-gram &
  {\color[HTML]{e38052} {\ul \textit{0.97}}} &
  {\color[HTML]{e38052} {\ul \textit{0.94}}} &
  {\color[HTML]{e38052} {\ul \textit{0.92}}} \\
 &
   
  7-gram &
  {\color[HTML]{e38052} {\ul \textit{0.99}}} &
  {\color[HTML]{e38052} {\ul \textit{0.97}}} &
  {\color[HTML]{e38052} {\ul \textit{0.94}}} \\ \bottomrule
\end{tabular}
}
\caption{The Spearman correlation values for all measures. Significance levels: \textit{0.05}, \textit{\underline{0.025}}, \textbf{\textit{\underline{0.01}}}, \color[HTML]{3353b7}{\textbf{\textit{\underline{0.005}}}}, \color[HTML]{e38052}{\textbf{\textit{\underline{0.001}}}}.}
\label{tab:spearman}
\end{table}

Lastly, we take a look at the Top-N evaluation in Table \ref{tab:topn}. We see that metrics select the best model in 70\% of the cases for the Bullinger\textsubscript{825} test set and in 60\% of the cases for the Gwalther\textsubscript{57} test set, respectively. This is in stark contrast to the Gwalther\textsubscript{433} validation set, with only 30\% Top-1 hits. This is mainly because the CERs of the models are very close to each other on the validation set, thus making it harder for alternative metrics to pick the best model. All in all, our metrics select one of the five best performing models 90\% of the time for the Bullinger\textsubscript{825} set and  80\% for the Gwalther\textsubscript{57} set, thus confirming that our metrics are capable of model selection.

\begin{table}[ht]
\centering
\scalebox{0.65}{
\begin{tabular}{llccc}
\toprule
\multicolumn{2}{l}{\multirow{2}{*}{\textbf{Metrics}}}                                                   & \multicolumn{3}{c}{\textbf{Ranking reference}} \\
& & \multicolumn{1}{c}{Gwalther\textsubscript{433}} & \multicolumn{1}{c}{Bullinger\textsubscript{825}} & \multicolumn{1}{c}{Gwalther\textsubscript{57}} \\ \midrule
\multirow{2}{*}{PPPL}             & \multicolumn{1}{l|}{BERT} & 1              & 1             & 1             \\
                                            & \multicolumn{1}{l|}{RoBERTa}     & 1              & 1             & 1             \\ \midrule
PPL                               & \multicolumn{1}{l|}{Statistical LM}               &                & 3             &               \\ \midrule
Token ratio                    & \multicolumn{1}{l|}{} & 3              & 1             & 1             \\ \midrule
\multirow{6}{*}{Character \textit{n}-grams} &           \multicolumn{1}{r|}{2-gram}           &                &               &               \\
                                                                         & \multicolumn{1}{r|}{3-gram     }      &                & 3             & 5             \\
                                                                         & \multicolumn{1}{r|}{4-gram    }       &                & 1             & 3             \\
                                                                      & \multicolumn{1}{r|}{5-gram   }        & 5              & 1             & 1             \\
                                                                      & \multicolumn{1}{r|}{6-gram  }         & 1              & 1             & 1             \\
                                                                         & \multicolumn{1}{r|}{7-gram }          & 3              & 1             & 1             \\ \bottomrule
\end{tabular}
}
\caption{Top-N values for each metric.}
\label{tab:topn}
\end{table}

\section{Conclusion}
This paper proposes metrics based on lexical features and (pseudo-)perplexity scores for HTR quality estimation and model selection. We confirmed other research that the lexicality of an HTR result is a strong indicator of quality. However, lexical resources are not always readily available, especially for low-resourced languages. While we dealt with Latin in this paper, we would like to recall that the Bullinger correspondence contains Early New High German texts (ENHG). We would need to assemble a big reference corpus for our or other lexicon-based methods to work. Our experiments confirmed that HTR quality estimation metrics based on transformers like multilingual BERT are suitable replacements. We leave it for future work whether pseudo-perplexity aligns with CER for historical German. After all, ENHG is not part of multilingual BERT's training data, and it is far more irregular and includes many more spelling variants (especially compared to Latin).

The pseudo-perplexity score also proved capable of ranking HTR outputs. Applying different models to new data and evaluating the output with ground truth-free metrics also opens up new possibilities to further improve HTR quality. For example, when we apply the metrics to lines of different model outputs, we can puzzle together the lines with the best scores, thus combining the best possible outputs from different models.

Our findings are valuable for digitisation initiatives by libraries and archives, which are concerned with the automatic transcription of handwritten documents. Especially our suggested employment of transformers works out of the box and provides quality estimates, based on which rankings of different models are possible. Hence, libraries and archives could quickly produce estimates of the HTR (or OCR) and identify quality issues. It needs further research to investigate whether such metrics are suitable for mass application.

\section{Acknowledgements}
We express our gratitude to the UZH Foundation and the various sponsors of
this project\footnote{see \url{https://www.bullinger-digital.ch/about}}. We thank the Zurich State Archive and
the Zurich Central Library for providing the scans.

\section{Bibliographical References}\label{reference}

\bibliographystyle{lrec2022-bib}
\bibliography{lrec2022-htreval}

\end{document}